**Adaptive Feedforward Gradient Estimation in Neural ODEs**

Jaouad DABOUNOU*

1* Professor, Hassan First University of Settat, Morocco

*Corresponding author E-mail: jaouad.dabounou@uhp.ac.ma

**Abstract**

Neural Ordinary Differential Equations (Neural ODEs) represent a significant breakthrough in deep learning, promising to bridge the gap between machine learning and the rich theoretical frameworks developed in various mathematical fields over centuries. In this work, we propose a novel approach that leverages adaptive feedforward gradient estimation to improve the efficiency, consistency, and interpretability of Neural ODEs. Our method eliminates the need for backpropagation and the adjoint method, reducing computational overhead and memory usage while maintaining accuracy. The proposed approach has been validated through practical applications, and showed good performance relative to Neural ODEs state of the art methods.

## 1. Introduction

Residual Networks (ResNet) was introduced in 2015, and helped to eliminate the problems posed by very deep networks [1]. The hidden states in a ResNet evolve using the following expression:

$$\mathbf{y}_{t+1} = \mathbf{y}_t + \mathbf{f}(\mathbf{y}_t, \boldsymbol{\theta}_t) \tag{1}$$

By conceptually extending the number of hidden layers to infinity, the discrete step size becomes infinitesimal, leading to the following continuous formulation:

$$\frac{\mathrm{dy}(t)}{\mathrm{d}t} = \mathbf{f}(\mathbf{y}(t), t; \boldsymbol{\theta}) \tag{2}$$

$$\mathbf{y}(0) = \mathbf{X}_0$$

where $\mathbf{X}_0$ represents the input layer.

As proposed in the paper "Neural Ordinary Differential Equations" [2], we can view the hidden states $\mathbf{y}(t)$ as evolving within a continuous domain, rather than through a discrete sequence of layers. The dynamics of this evolution are governed by the function $\mathbf{f}$, which depends on the current state $\mathbf{y}(t)$ at time $t$, and the parameters $\boldsymbol{\theta}$ of the "continuous" network.

Given that $\mathbf{y}(0)$ is known (corresponding to the input data), forward propagation is equivalent to solving an ordinary differential equation (ODE).

To obtain the network's output, we integrate this ODE from $t$=0 to $t$=$T$. The analytical representation of this integration is:

$$\mathbf{y}(T) = \mathbf{y}(0) + \int_0^T \mathbf{f}(\mathbf{y}(t), t; \boldsymbol{\theta}(t))\mathrm{d}t \tag{3}$$

In practice, the integral (3) is typically computed using numerical integration methods. The simplest such method is the Euler method with step size $\Delta t$=1, which when applied here, corresponds to a ResNet:

$$\mathbf{y}(t+\Delta t) = \mathbf{y}(t) + \Delta t\, \mathbf{f}(\mathbf{y}(t), t; \boldsymbol{\theta}) \tag{4}$$

However, most neural ODE networks employ more sophisticated ODE solvers that offer superior efficiency and accuracy, although at the cost of increased complexity. We then see that the integration of the ODE constitutes the core of the necessary calculations to be carried out, however, in scientific literature it is stated that the output of the network can be computed by a black-box differential equation solver [2,3,4], which evaluates the hidden unit dynamics $\mathbf{f}(\mathbf{y}, t; \boldsymbol{\theta})$ wherever necessary to determine the solution with the desired accuracy. The output, $\mathbf{y}(T)$ is then used to calculate the loss function $L(\mathbf{y}(T))$. And to train the neural network, we calculate the gradients:

$$\frac{d\mathbf{y}(T)}{d\boldsymbol{\theta}} \text{ and } \frac{dL(t)}{d\boldsymbol{\theta}} \tag{5}$$

with respect to the parameters $\boldsymbol{\theta}$.

The backpropagation can be achieved through the operations of the solver, especially facilitated by automatic derivation. But, given the complexity of the algorithms used in the forward integration, making a derivation in the reverse path would be expensive in terms of calculation cost and memory. So, as if to acknowledge the fact that the solver is a black box, researchers propose to construct a new ordinary differential equation using the adjoint method [2,3,4,5].

The key idea of this approach is to define an adjoint variable $\mathbf{a}(t)$ that encapsulates information on how the loss $L$ changes in response to infinitesimal changes in the state $\mathbf{y}(t)$ over time. The adjoint variable is therefore defined as follows:

$$\mathbf{a}(t) = \frac{\partial L}{\partial \mathbf{y}(t)} \tag{6}$$

To calculate the sensitivity of $L$, accumulated over the entire interval [0,$T$], we must integrate the equation $\frac{d\mathbf{a}(t)}{dt}$ over this interval. This integration is done from $T$ to 0 with the initial condition $\mathbf{a}(T)$ which we can calculate directly as:

$$\mathbf{a}(T) = \frac{dL}{d\mathrm{y}(T)} \tag{7}$$

The equation to integrate is essentially written:

$$\frac{d\mathbf{a}(t)}{dt} = -\mathbf{a}(t)^{\mathrm{T}} \frac{\partial \mathbf{f}(\mathbf{y}(t),t;\boldsymbol{\theta})}{\partial \mathbf{y}(t)} \quad (8)$$

In this work, we propose an approach that consists in evolving the EDO during the forward pass by combining both the integration of the equation of the dynamics of hidden states, at the same time as that of the sensitivity of these hidden states to the parameters of the model. We will show that this approach allowed for bypassing the need to backpropagate gradients through the complex ODE solver used in the forward pass while ensuring consistency between the calculation of hidden states and the estimation of the gradients of the loss function [2 to 5].

## 2. ODE integration approach

For integrating a system of ordinary differential equations

$$\frac{d\mathbf{y}(t)}{dt} = \mathbf{f}\big(t, \mathbf{y}(t)\big) \quad (9)$$

we consider here the fourth order Runge-Kutta method [6], which is the most often used scheme in practice. The solution is advanced through a subset $[t_n, t_{n+1}]$, $t_n = t_n + h$, using the following formulas:

$$\mathbf{k}_1 = \mathbf{f}(t_n , \mathbf{y}_n)$$

$$\mathbf{k}_2 = \mathbf{f}(t_n + \frac{h}{2} , \mathbf{y}_n + \frac{h}{2}\mathbf{k}_1)$$

$$\mathbf{k}_3 = \mathbf{f}(t_n + \frac{h}{2} , \mathbf{y}_n + \frac{h}{2}\mathbf{k}_2) \quad (10)$$

$$\mathbf{k}_4 = \mathbf{f}(t_n + h , \mathbf{y}_n + h\, \mathbf{k}_3)$$

and

$$\mathbf{y}_{n+1} = \mathbf{y}_n + \frac{h}{6}(\mathbf{k}_1 + 2\,(\mathbf{k}_2 + \mathbf{k}_3) + \mathbf{k}_4) \quad (11)$$

Then

$$\mathbf{y}_{n+1} = \mathbf{y}(t_{n+1}) + O(h^5) \quad (12)$$

where we suppose that $\mathbf{y}_n = \mathbf{y}(t_n)$.

A commonly employed technique for stepsize control in this method is step doubling [6,7]. However, this approach necessitates 11 derivative evaluations per step, often requiring substantial computing time.

In the subsequent section, we present a novel stepsize control technique that proves to be computationally more efficient than step doubling. These mathematical advancements draw inspiration from our prior research on structure modeling, where a similar Runge-Kutta integration method was utilized [8]. However, their application to deep learning represents a new and innovative direction.

## 3. Zero-Cost Step Adaptation Runge-Kutta Method

Runge-Kutta algorithms are largely used to integrate Ordinary Differential Equations, thanks to their precision and the stability of the calculations they offer despite being explicit methods. In addition, they provide interesting possibilities for predicting the size of integration steps [6,7,8].

Inspired by methods used at early 1990 in the integration of the differential equations of elastoplasticity and structural equilibrium [8], we propose a technique that enables the prediction of the integration step size without additional cost. Furthermore, the right-hand side of the differential equation $\mathbf{f}(t, \mathbf{y}(t))$, will be approximated over the interval $[t_n, t_n+h]$ by second-degree polynomial $\mathbf{g}(t)$ with an approximation error of $O(h^3)$. The approximation of $\mathbf{f}$ over the entire interval $[0, T]$, which is equal to $\mathbf{g}(t)$ over each sub-interval will be continuous and twice differentiable at the boundaries of the intervals.

Similarly, the solution, the solution $\mathbf{y}(t)$ will be approximated on the subinterval $[t_n, t_n+h]$ by a by a third-degree polynomial $\boldsymbol{\phi}(t)$ with an approximation error of $O(h^4)$. The polynomial $\boldsymbol{\phi}(t)$ is computed from $\mathbf{g}(t)$ by simple integration, ensuring that it extends over $[0, T]$ and presents the resulting regularities (see Appendix for details).

In the approach we present, let $\mathbf{k}_1$, $\mathbf{k}_2$, $\mathbf{k}_3$ and $\mathbf{k}_4$ be approximate derivative values given by equation (10). The function $\mathbf{f}$ is approximated over $[t_n, t_{n+1}]$ by $\mathbf{g}$, the interpolating polynomial such that:

$$\mathbf{g}(t_n) = \mathbf{k}_1$$

$$\frac{\mathrm{d}\mathbf{g}}{\mathrm{d}t}(t_n) = \frac{1}{h}(-3\mathbf{k}_1 + 2(\mathbf{k}_2 + \mathbf{k}_3) - \mathbf{k}_4)$$

$$\mathbf{g}(t_n + \tfrac{h}{2}) = \tfrac{1}{2}(\mathbf{k}_2 + \mathbf{k}_3) \tag{13}$$

$$\mathbf{g}(t_n + h) = \mathbf{k}_4$$

The expression of $\mathbf{g}(t)$ can be written**:**

$$\mathbf{g}(t) = \mathbf{a}_0 + \frac{\mathbf{a}_1}{h}(t_n)\,(t - t_n) + \frac{\mathbf{a}_2}{h^2}(t - t_n)^2 \tag{14}$$

where

$\mathbf{a}_0 = \mathbf{k}_1$

$\mathbf{a}_1 = -3\,\mathbf{k}_1 + 2\,(\mathbf{k}_2 + \mathbf{k}_3) - \mathbf{k}_4$ (15)

$\mathbf{a}_2 = 2\,(\mathbf{k}_1 - (\mathbf{k}_2 + \mathbf{k}_3) + \mathbf{k}_4)$

**g** satisfies (Appendix) the following relation:

$$\mathbf{g}(t) = \mathbf{f}\big(t\,,\mathbf{y}(t)\big) + \mathrm{O}(h^3) \tag{16}$$

Furthermore, if $\tau = \frac{t - t_n}{h}$, then one has

$$\mathbf{g}(t) = \mathbf{p}(\tau) = \mathbf{a}_2\tau^2 + \mathbf{a}_1\tau + \mathbf{a}_0 \tag{17}$$

Where $\mathbf{a}_0$, $\mathbf{a}_1$ and $\mathbf{a}_2$ are given by relation (15), and the integral

$$\mathbf{q}(\tau) = \mathbf{y}_n + \int_0^{\theta} h\mathbf{p}(\xi)d\xi = \mathbf{z}_n + \int_{t_n}^{t} \bar{\mathbf{f}}(\tau)d\tau \tag{18}$$

so that

$$\mathbf{q}(\tau) = \mathbf{y}_n + h\tau\left(\frac{1}{3}\mathbf{a}_2\tau^2 + \frac{1}{2}\mathbf{a}_1\tau + \mathbf{a}_0\right) \tag{19}$$

satisfies for $\tau = 1$, the equation

$$\mathbf{q}(1) = \mathbf{y}_{\boldsymbol{n}} + \frac{h}{6}(\mathbf{k}_1 + 2(\mathbf{k}_2 + \mathbf{k}_3) + \mathbf{k}_4) \tag{20}$$

which coincides with relation (11).

Equations (12) and (17) enable us to estimate the truncation error of the approximation (16). Specifically, the error can be expressed as:

$$\mathbf{e}(1) = \mathbf{p}(1) - \mathbf{f}(t_n+h, \mathbf{q}(1)) = \mathbf{p}(1) - \mathbf{f}(t_n+h, \mathbf{y}(t_n+h)) + \mathrm{O}(h^5) \tag{19}$$

The cost of calculating the error in the proposed scheme can be considered negligible, as it only requires the calculation of $\mathbf{f}(t_n+h, \mathbf{y}(t_n+h))$, which will be used as $\mathbf{k}_1$ in the next integration step. Of course, this calculation will not be utilized if the integration on the sub-step fails, but efforts are made to ensure that such failures are rare.

## Application

We will show on simple examples the self-adaptive nature of the method presented. We propose to integrate the following differential equation:

Consider over [0,10] the ODE

$$\begin{cases} f'(t) = \cos(t) + \dfrac{(2t-3-7)\exp\big(-(t-3)(t-7)\big)}{1+\exp\big(-(t-3)(t-7)\big)} \\ f(0) = 1 \end{cases}$$

The analytical solution of this EDO is given by:

$$f(t) = \sin(t) + \frac{1}{\big(1+\exp\big(-(t-3)(t-7)\big)\big)}$$

We choose the initial substep $h_0 = 0.1$, and the required accuracy $\varepsilon = 1e^{-2}$.

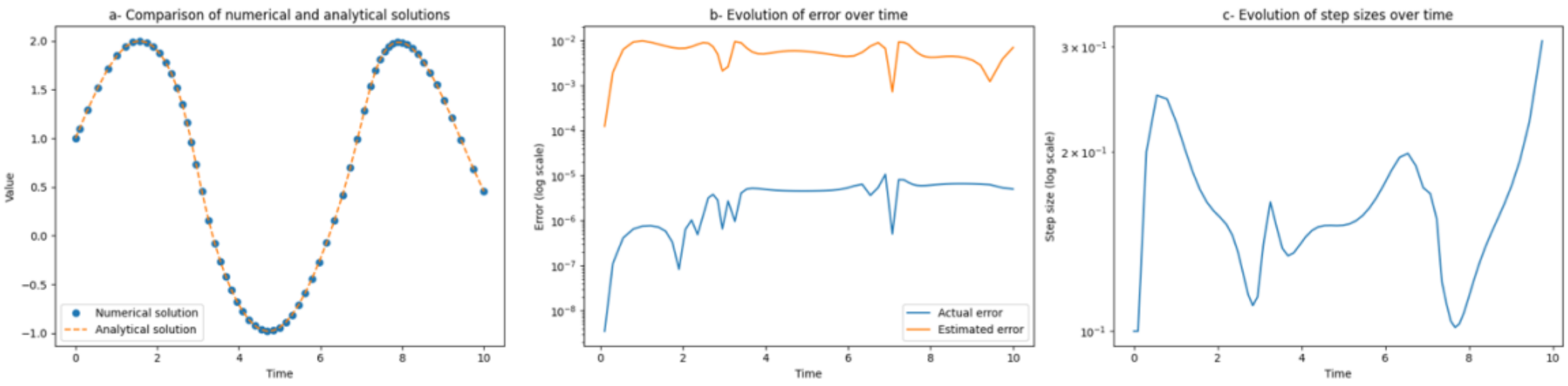

Figure 1. Self-adaptivity of the integration scheme to nonlinearities

In the second example, we consider a function that is not differentiable, even if we have given an artificial value to the derivative to let the algorithm do its thing. Figure 1b and Fig 1.b show that the algorithm reduces the size of the steps to cross the singular points, and increases in regular areas, in a way to respect the required precision.

Consider over [0, 2π] the ODE

$$\begin{cases} f'(t) = \begin{cases} -\sin(t) \;\; \text{if } 0 \le t \le \dfrac{3\pi}{4 \text{ or } 5\pi 4} \le t \le 2\pi \\ 0 \;\; \text{if } \dfrac{3\pi}{4} < t < \dfrac{5\pi}{4} \end{cases} \\ f(0) = 1 \end{cases}$$

The analytical solution of this EDO is given by:

$$
f(t) = \begin{cases} \cos(t) \quad \text{if } 0 \le t \le \dfrac{3\pi}{4 \text{ or } 5\pi 4} \le t \le 2\pi \\ \cos\left(\dfrac{3\pi}{4}\right) \text{if } \dfrac{3\pi}{4} < t < \dfrac{5\pi}{4} \end{cases}
$$

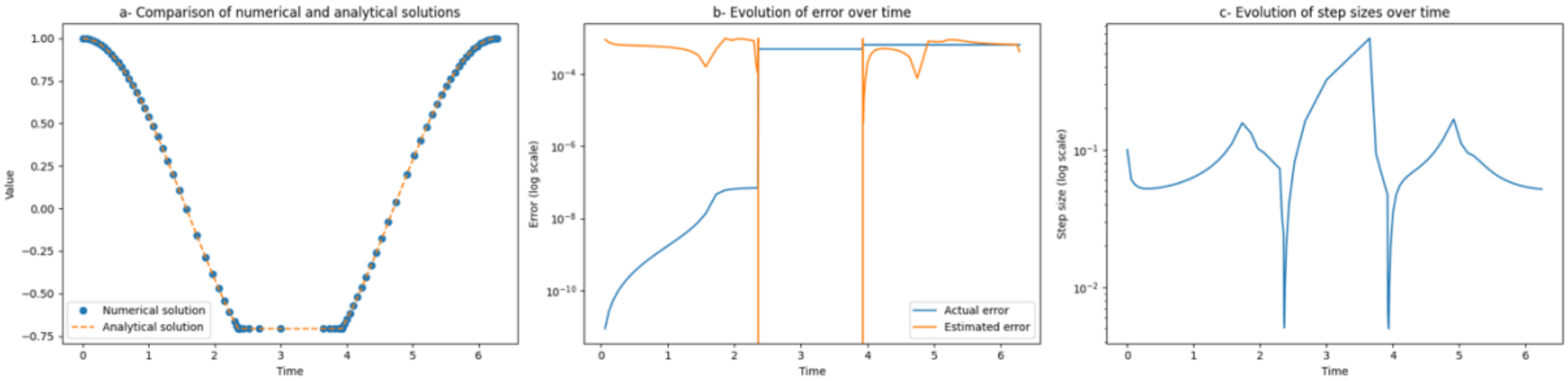

Figure 2. Self-adaptivity of the integration scheme to nonlinearities

## Application to chaotic systems

We also applied our approach to the Van de Pol and Lorentz equations. The figures show how the pitch is adapted to the non-linearities encountered. We notice periodic adaptation, and a stability in face of the periodic behavior of the chaotic system.

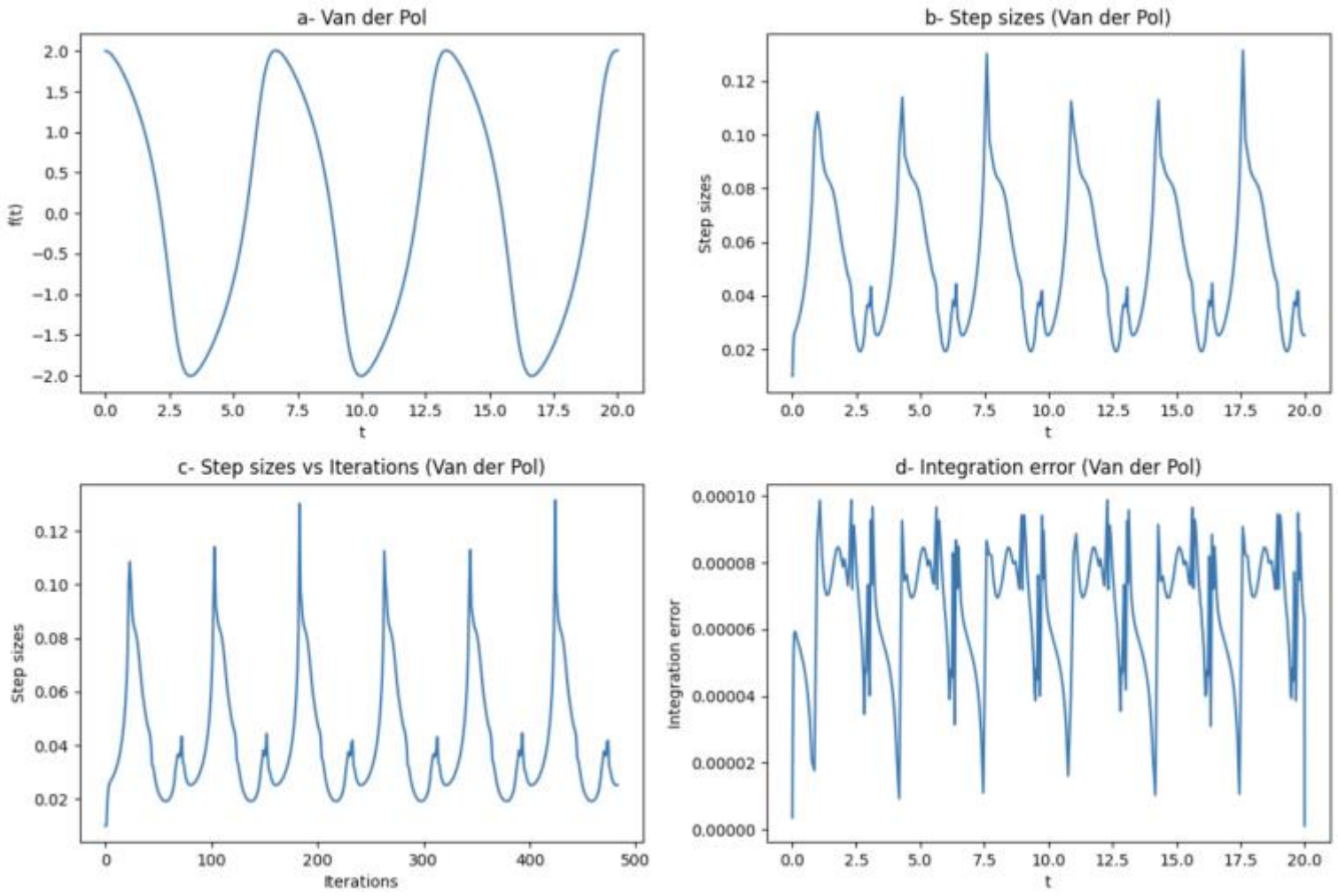

Figure 3. Van der Pol – Stability of the self-adaptive pattern and technique

Figure 3. shows And for the Lorentz equation that the algorithm can adapt to complexities of Lorenz equation, and adapt the steps to maintain the error verifying the required accuracy.

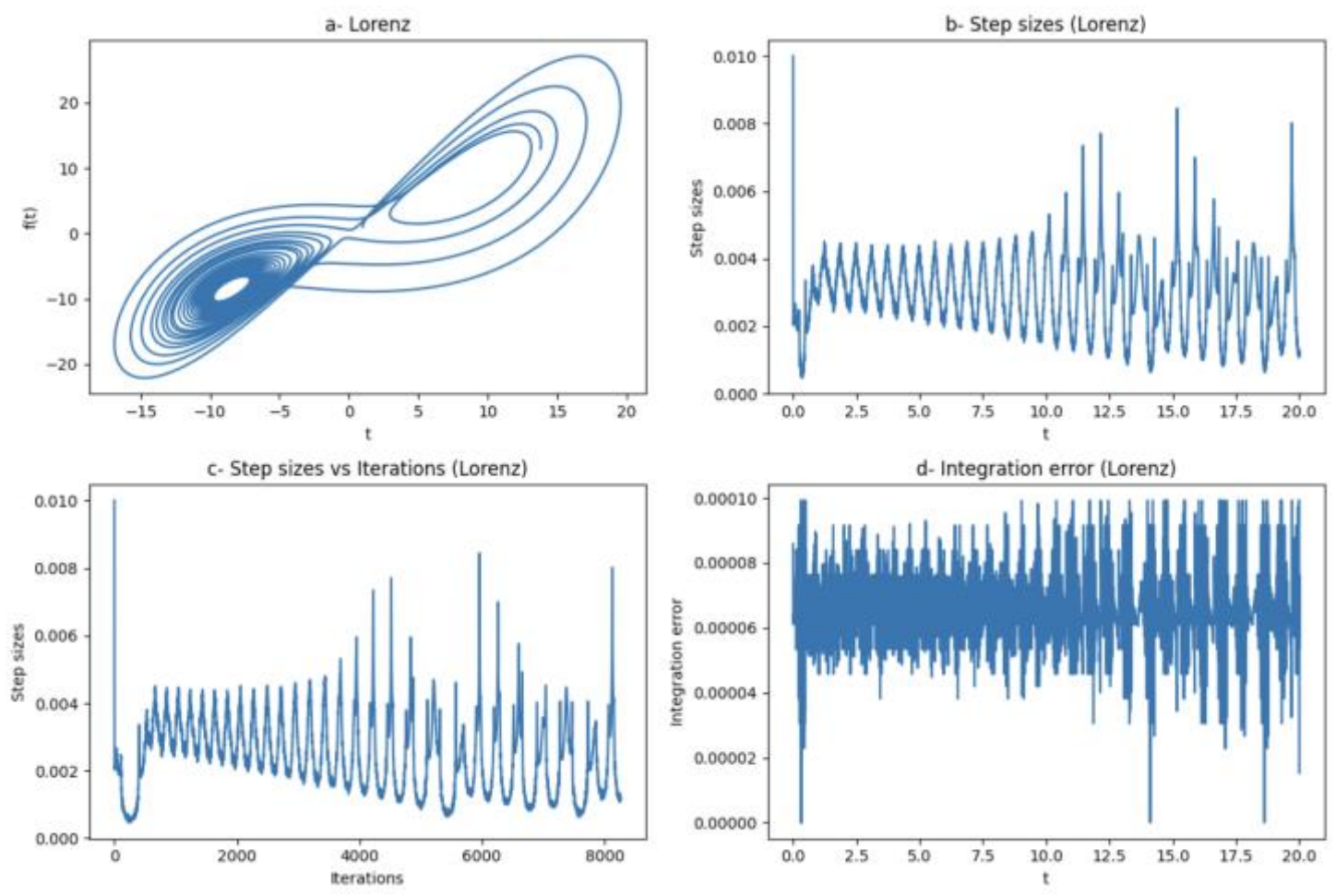

Figure 4. Lorenz - Self-Adaptivity to Bumpy Systems

## 4. A feedforward consistent gradient estimation

### 4.1. No Adjoint Method, no gradient Backpropagation:

The ODE solver advances the hidden state from $t_n$ to $t_{n+1} = t_n + h$, by integrating the ODE in relation (2). The introduction of the adjoint method was initially considered a key contribution of the Neural ODEs paper and subsequent works. It was particularly emphasized that this approach allowed for bypassing the need to backpropagate gradients through the complex ODE solver used in the forward pass [2 to 5].

However, it's crucial to remember that gradients should be consistent with the actual computations performed by the network. Therefore, we should focus on the approximations used to compute the output in the forward pass, not the underlying differential equation itself. We combine, in the forward pass, both the calculation of the hidden states and the estimation of related gradients to model parameters.

Our method builds upon the Neural ODE framework, which formulates the evolution of a neural network's hidden states as the solution to an ODE:

$$\frac{d\mathbf{y}(t)}{dt} = \mathbf{f}(t, \mathbf{y}(t);\ \boldsymbol{\theta})$$

We augment this equation with an additional term that captures the derivative of the state with respect to the parameters:

$$\frac{d}{dt}\left(\frac{d\mathbf{y}(t)}{d\boldsymbol{\theta}}\right) = \frac{d}{d\boldsymbol{\theta}}\mathbf{f}(t, \mathbf{y}(t);\ \boldsymbol{\theta})$$

This augmented ODE is integrated using the RK4 method with adaptive step size control, ensuring efficient and accurate computation.

For the augmented state, the updates are:

$$\left[\frac{d\mathbf{y}(t)}{dt}, \frac{d}{dt}\left(\frac{d\mathbf{y}(t)}{d\boldsymbol{\theta}}\right)\right] = \left[\mathbf{f}(t, \mathbf{y}(t);\ \boldsymbol{\theta}), \frac{d\mathbf{f}(t, \mathbf{y}(t);\ \boldsymbol{\theta})}{d\boldsymbol{\theta}}\right]$$

At $t$=1, we obtain both $\mathbf{y}(1)$ and $\frac{d\mathbf{y}(1)}{d\boldsymbol{\theta}}$, which allows us to compute the gradient of the loss function $L$ without full backpropagation.

**4.2. Gradient of the Loss Function**

The loss function $L$ is typically defined as the cross-entropy loss between the predicted outputs and the true labels. The gradient of the loss with respect to the parameters $\boldsymbol{\theta}$ is computed as follows:

$$\frac{dL}{d\boldsymbol{\theta}} = \frac{\partial L}{\partial \mathbf{y}(1)} \frac{d\mathbf{y}(1)}{d\boldsymbol{\theta}}$$

Instead of using the traditional backpropagation [9,10,11], which requires storing all intermediate states, or using Adjoint method [2, 12, 13], which may introduce inconsistencies between the forward and backward passes, we calculate and aggregate gradients in real-time during the integration process. Specifically, the term $\frac{d\mathbf{y}(t)}{d\boldsymbol{\theta}}$ is incrementally updated at each integration step, allowing us to obtain $\frac{d\mathbf{y}(1)}{d\boldsymbol{\theta}}$, and the Loss gradient $\frac{dL}{d\boldsymbol{\theta}} = \frac{\partial L}{\partial \mathbf{y}(1)} \frac{d\mathbf{y}(1)}{d\boldsymbol{\theta}}$ at $t$=1 without storing intermediate activations. The gradient accumulation is conditioned on the integration error being within a predefined tolerance.

**4.3. Stability and convergence of the augmented method**

The stability and convergence analysis of the standard Runge-Kutta method is based on two key criteria: consistency and stability. Convergence arises from the fact that, if a method is both consistent and stable, it is convergent.

**Consistency**

The Runge-Kutta method of order 4 is consistent because the local truncation error is of order $O(h^5)$, which means that the smaller the integration step $h$, the faster the error decreases. The consistency of the method lies in its ability to correctly approximate the derivative of a system of differential equations with an accuracy of order 4.

In the context of our augmented approach, we integrate both the hidden states equation (ODE) and the equation of state derivatives with respect to the $\boldsymbol{\theta}$ parameters, thus forming a coupled system of equations. The consistency of the method remains valid as long as the polynomial approximation used in both integrations satisfies the conditions of regularity (continuity of the partial derivatives of $\mathbf{f}$).

**Stability**

The numerical stability of the RK4 scheme is well known for non-stiff EDO systems. The addition of the augmented component (derivatives with respect to the parameters) does not affect the stability. In this case, the adaptive integration step we propose helps to ensure that the schema remains stable even in the presence of strong nonlinearities in the data or parameters.

**Convergence for the augmented method**

We can formalize the convergence of the augmented method by adapting the classical results on the convergence of Runge-Kutta methods. Consider an augmented differential equation. Convergence is ensured if the local truncation error in the two equations (hidden states and gradients with respect to the parameters) is of the order $O(h^5)$, and if the overall error is controlled by $O(h^4)$.

**Impact on generalization**

The impact of convergence and stability on generalization is related to the model's ability to avoid overfitting noisy data. The use of high-order polynomials to approximate the evolution of hidden states (and their derivatives) introduces a form of natural regularization (Relations (17) and (19)). This bias in favor of regular functions improves generalization by avoiding learning from rapid fluctuations or noise in the data. It also allows you to check the capacity of the model, since polynomial approximations impose a certain softness on the solutions.

## 5. Experiments

### 5.1. Image Classification on MNIST

This experiment compares the proposed Adaptive Feedforward Gradient Estimation (AFGE) approach to the standard Neural ODEs approach presented in [2] for image classification using the MNIST dataset. We analyze the performance in terms of test accuracy and training time over 10 epochs. The results demonstrate that the proposed approach achieves slight better accuracy with significant improved training efficiency.

Results:

1. Test Accuracy: Figure 5 shows that both approaches achieve high accuracy, consistently above 98%. The proposed approach reaches a peak accuracy of 99.41% at epoch 10, while the classical approach peaks at 99.32% at epoch 7. Overall, the accuracy trajectories are similar, with slight variations between epochs.

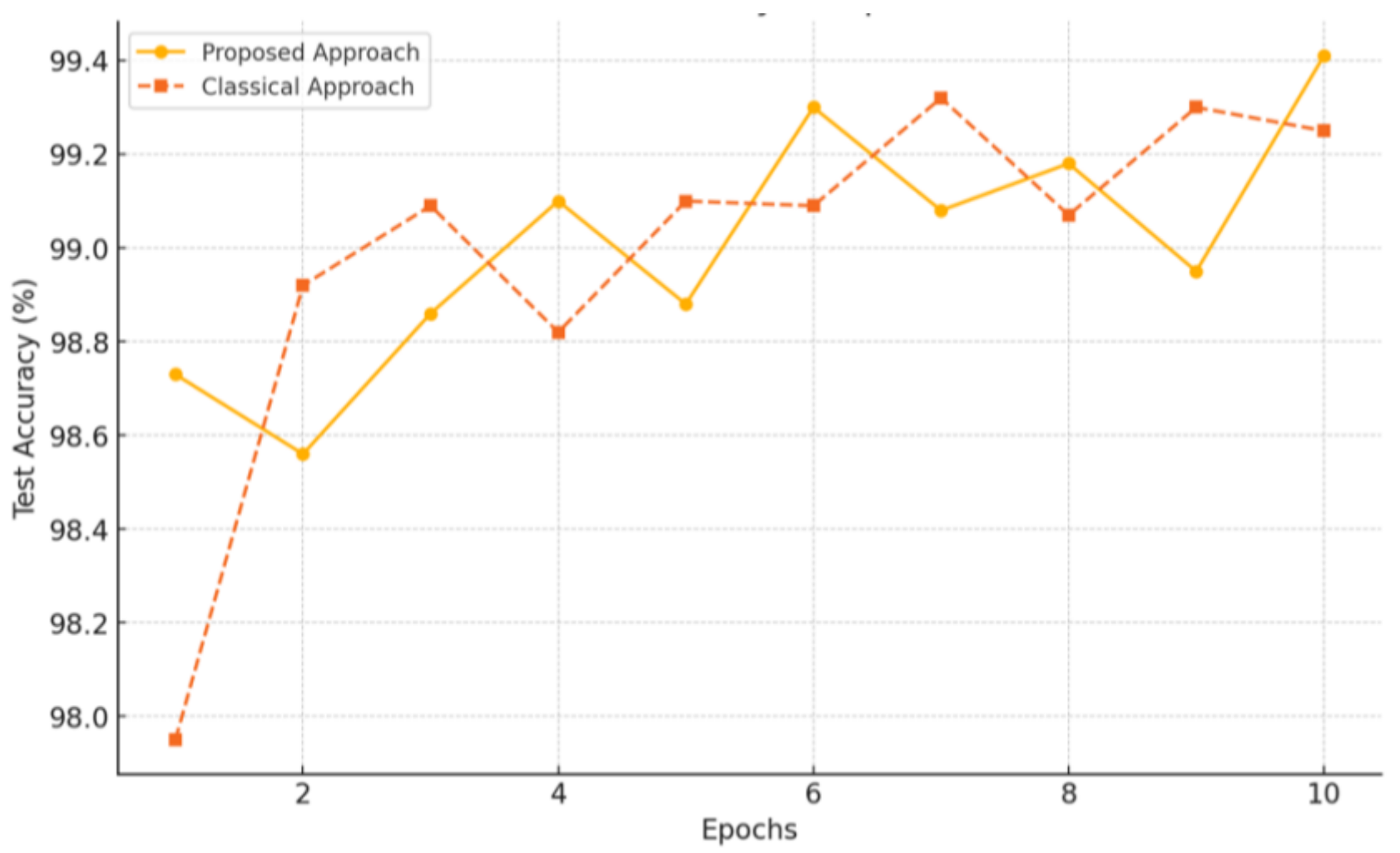

Figure 5. Evolution of test loss over 10 epochs

2. Training Time: Figure 6 shows that the proposed approach demonstrates consistently lower training times across most epochs. The average training time per epoch for the proposed approach is approximately 67.38 seconds, compared to 76.65 seconds for the classical approach when excluding the anomalous epoch 6 for the classical approach, and 106.65 seconds for the classical approach when including all epochs. This shows an efficiency gain in the proposed method. This efficiency gain is particularly pronounced in epochs where the classical approach encounters computational bottlenecks and shows considerable instability in training time as at epoch 6 in Figure 6.

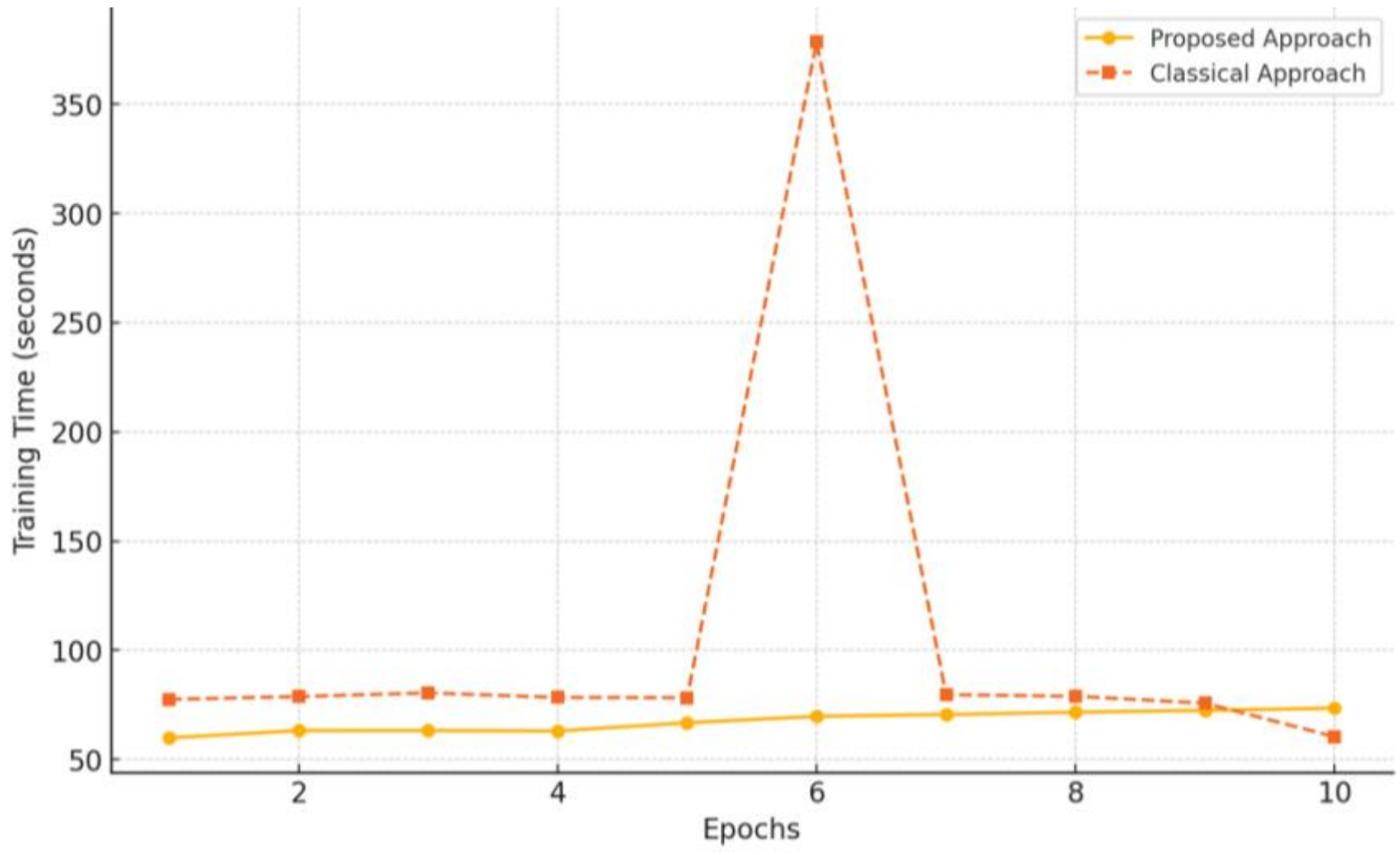

Figure 6. Evolution of time training over 10 epochs

## 5.2.Validation on CIFAR-10 dataset

Dataset: CIFAR-10 dataset, consisting of 60,000 32x32 color images in 10 classes (50,000 for training and 10,000 for testing).

Models: Both approaches used a ResNet-18 architecture as the backbone, followed by an ODE layer and a final classification layer. The ODE function was integrated using the RK4 method with adaptive step size control.

Results:

- Test Accuracy:
    - The proposed method showed a more stable and consistent improvement in accuracy (Figure 7). The standard approach exhibited more variability between epochs.
    - By the final epoch, the proposed method achieved 81.71% accuracy, while the classical approach reached 80.27%.
- Training Time:
    - The proposed method consistently demonstrated shorter training times per epoch compared to the classical approach (Figure 8).
    - On average, the proposed method reduced training time by approximately 40% per epoch.
- Performance Analysis:
    - The AFGE's enhanced performance can be attributed to improved gradient flow through the ODE layer, allowing for more effective learning of complex feature representations.
    - The significant decrease in computational time for AFGE is essentially due to the feedforward estimation of the gradients, instead of the Adjoint method used in the standard approach.
    - The learning curve of the proposed method displayed more stability and consistent improvement, suggesting faster convergence and a more robust, generalizable model.

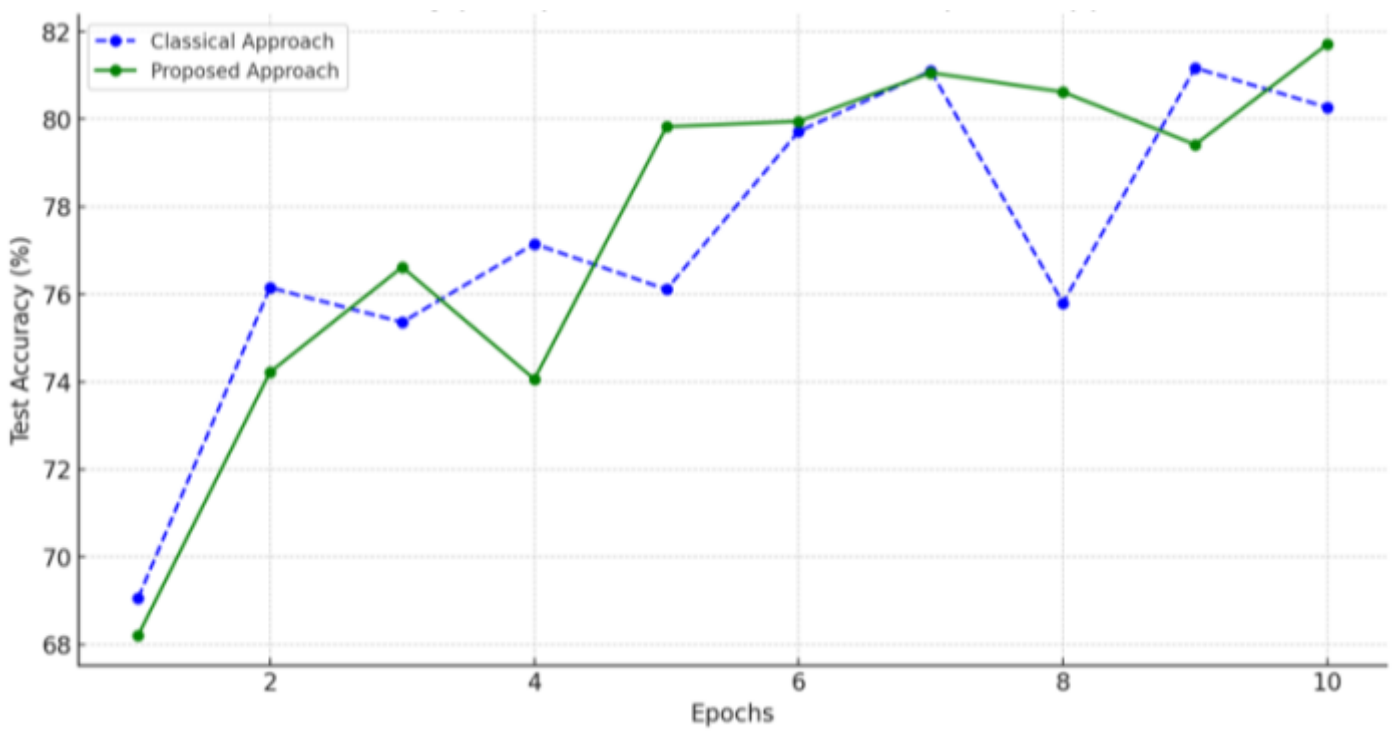

Figure 7. Evolution of time training over 10 epochs

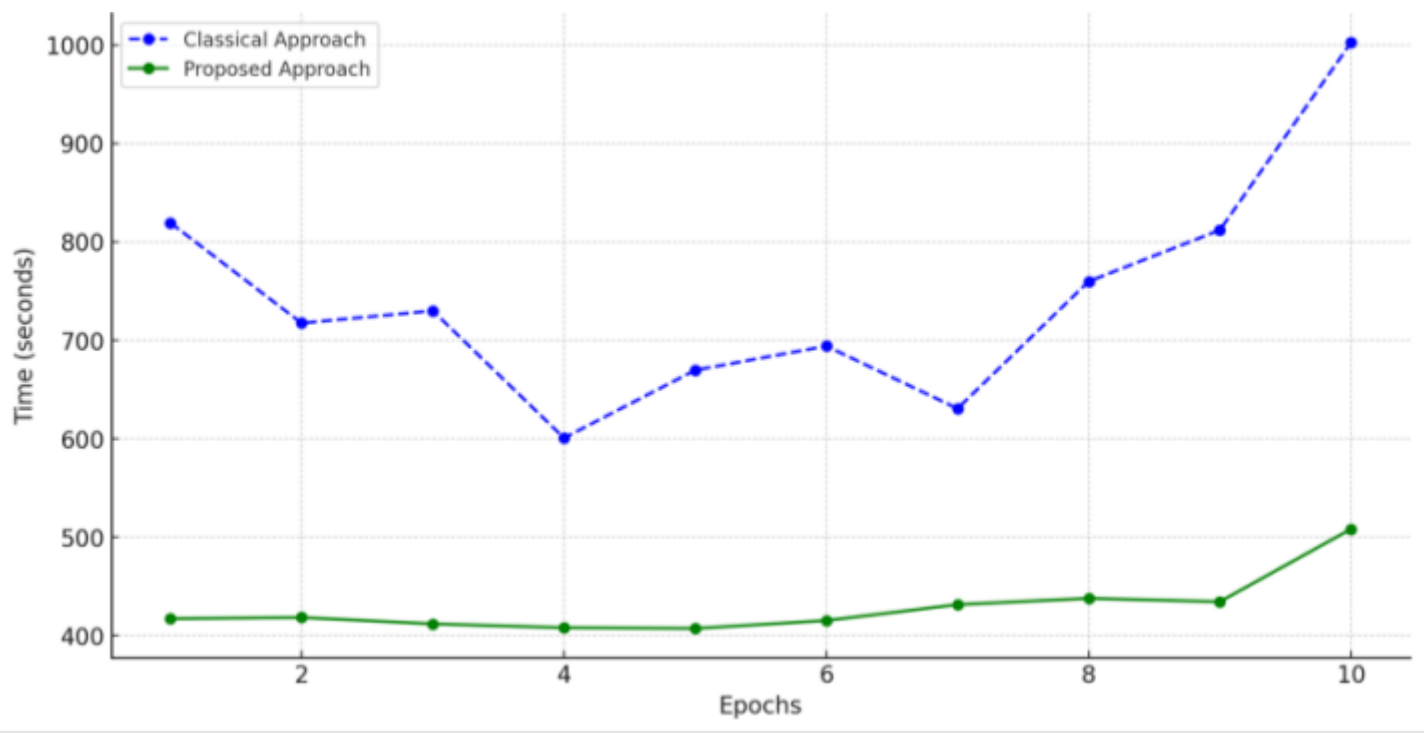

Figure 8. Evolution of time training over 10 epochs

## 1.1.Validation on Melanoma Skin Cancer Dataset

- Dataset: Melanoma Skin Cancer Dataset consisting of 10,000 images (9,600 for training and 1,000 for testing).

This dataset is from Kaggle: https://www.kaggle.com/datasets/hasnainjaved/melanoma-skin-cancer-dataset-of-10000-images

- Models: Both approaches used a ResNet-50 model. The classical method employed static one-hot encoded labels from the start, while the AFGE method progressively evolved the target labels from uniform distributions to one-hot encodings over the course of training.

Results :

- Test Accuracy:
    - AFGE showed a generally higher and more stable test accuracy throughout the training process (Figure 9).
    - The Standard NODE Approach exhibited more fluctuations in accuracy.
    - By the 10th epoch, AFGE achieved 44.22% accuracy, while the Standard NODE Approach reached 44.07%.

- **Training Time**:
    - AFGE consistently demonstrated shorter training times per epoch compared to the Standard NODE Approach (Figure 10).
    - AFGE's training time increased gradually, starting at 489.14 seconds in the first epoch and reaching 747.80 seconds by the 10th epoch.
    - The Standard NODE Approach showed longer and more variable training times, ranging from 595.30 seconds in the first epoch to 1014.30 seconds in the 10th epoch.
    - Using the AFGE approach resulted in more than 20% reduction in total training time over 10 epochs compared to the Standard NODE approach.

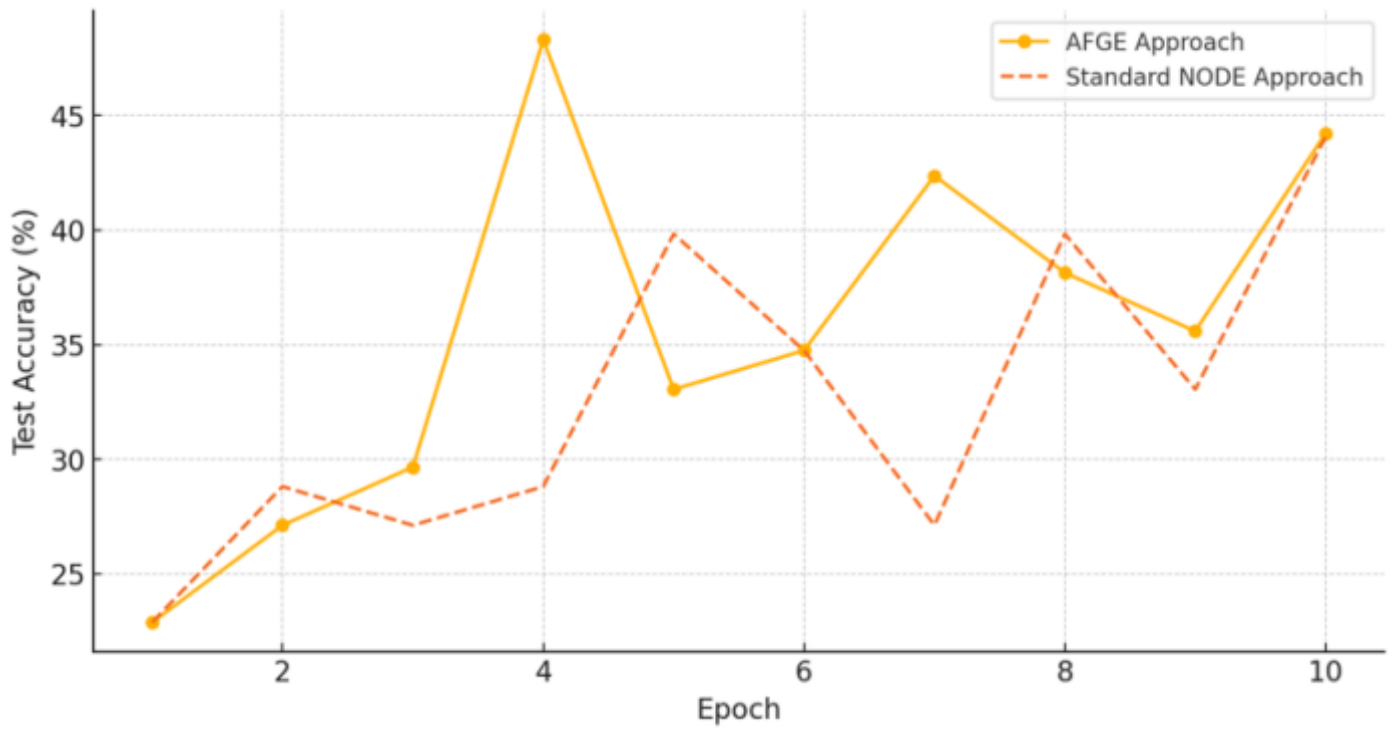

Figure 9: Test accuracy per epoch for AFGE and Standard NODE approaches on Melanoma Dataset

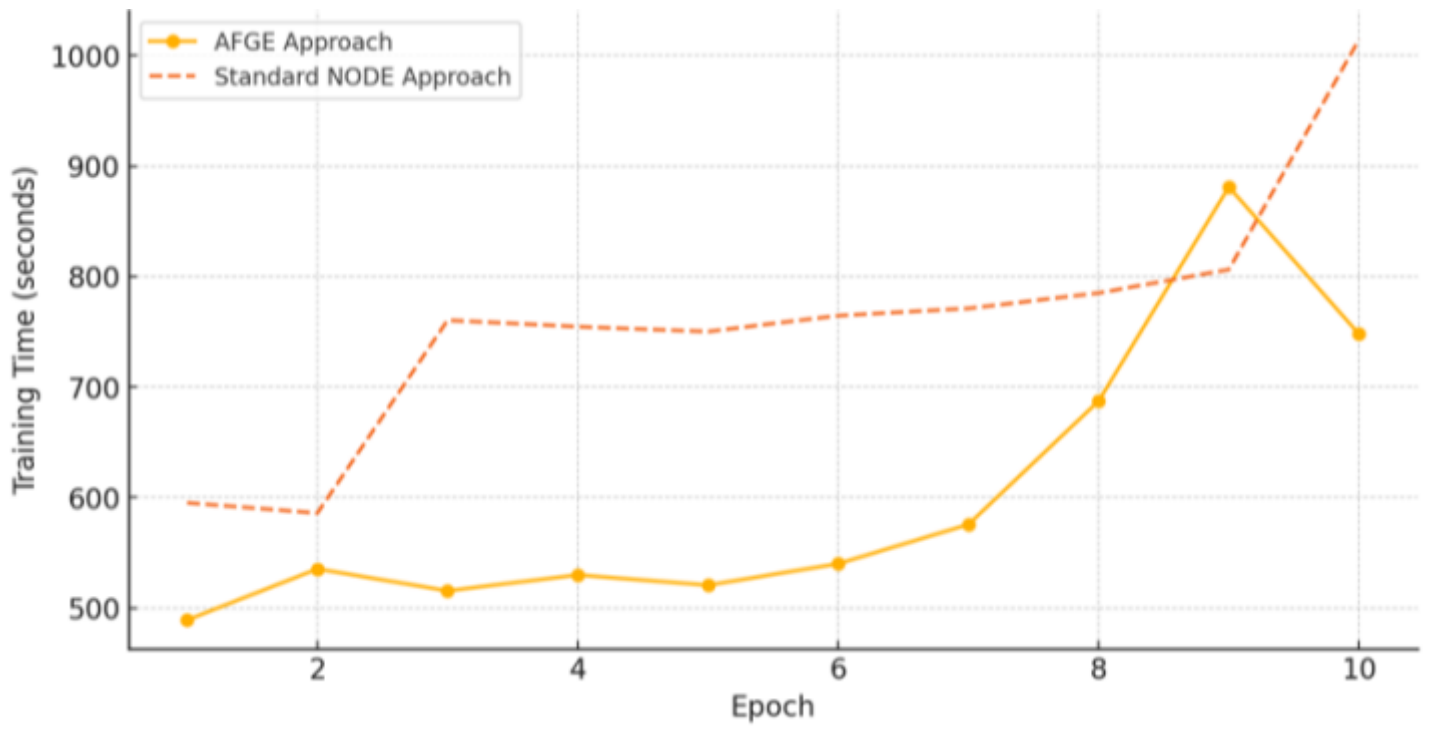

Figure 10: Training time per epoch for AFGE and Standard NODE approaches on Melanoma Dataset

## 6. Discussion

### 6.1. Advantages of the Proposed Approach

Standard Neural ODEs rely on full backpropagation through time, which can be computationally expensive and memory-intensive. Our proposed method, by augmenting the ODE with parameter derivatives, computes gradients more efficiently. This is particularly beneficial for deep networks and large datasets where computational savings can be substantial.

**Zero-Cost Step Adaptation:** Our approach, coupled with the Runge-Kutta method, introduces a zero-cost step adaptation mechanism. Local error estimation allows us to dynamically adjust the integration step size, and implicitly the number of layers in the neural network, based on the encountered nonlinearities, complexities, and the data being processed. This dynamic adaptation improves accuracy in complex regions, enhances computational efficiency in smoother areas, and implicitly adjusts the neural network's depth to the problem's complexity, optimizing resource allocation.

- **Continuous Gradient Propagation:** Computing gradients during the forward pass is a key contribution of this work. While applied here in the context of Neural ODEs, this approach can be generalized to other neural network architectures. It addresses many challenges associated with traditional backpropagation, including vanishing gradients, high computational demands, and memory constraints.

  Traditional backpropagation and the adjoint method require storing all intermediate states, which can be computationally expensive and memory-intensive, especially for large-scale problems. Our incremental gradient accumulation method tackles this by computing and aggregating gradients in real-time during integration.

  Moreover, our method provides a more consistent and accurate gradient estimation. By using polynomial approximations of hidden states and their derivatives, we ensure gradients are computed consistently with the forward integration process. This consistency is crucial for the stability and convergence of training.

- **Regularization and Overfitting Risk Reduction:** Representing hidden state dynamics with polynomial approximations introduces a natural form of regularization. By maintaining a polynomial form that is sufficiently close to the true dynamics yet regular enough, the model effectively reduces the risk of overfitting. This is particularly valuable when the model might capture noise in the data, thus improving generalization.

Approximating hidden state dynamics with polynomials introduces a bias towards simpler, more regular functions. This bias acts as regularization, similar to techniques like weight decay or dropout, but integrated directly into the model's architecture.

- **Transparency and Explainability:** Our approach also enhances transparency. The polynomial coefficients ($\mathbf{a}_0$, $\mathbf{a}_1$, $\mathbf{a}_2$) offer explicit insights into the evolution of hidden states over time. Each term in the polynomial has a clear interpretation, making the model's behavior more interpretable and providing a better understanding of the underlying processes within the ODE solver. This transparency is essential for building trust in Neural ODE models, especially in critical applications where model interpretability is paramount.

### 6.2. Limitations and Future Directions

This paper serves as an introduction to this new approach. We have deliberately chosen not to delve deeper into the various possibilities offered by this approach, which can be considered the main limitation of this work. For instance, we could further explore the coupling between the integration of hidden states and their gradients with respect to the parameters. We could also analyze potential simplifications within the specific context of the Runge-Kutta method. The polynomial approximations themselves offer great potential that remains unexplored in this introductory work. Limiting ourselves to the classification task is also a limitation. It's easy to see how the continuous progression of gradient estimation would be highly beneficial for time series. The polynomial expression of hidden states and gradients holds great potential in the field of model explainability, which we haven't explored in this introductory work.

## 2. Conclusion

In this paper, we introduced a novel approach to Neural ODEs that estimate continuously the gradient of the hidden states on the model's parameters during the feedforward pass. This approach enhances the efficiency, consistency, and interpretability of these models. By eliminating the need for backpropagation or the adjoint method, we reduce memory usage and computational overhead while ensuring that gradient calculations remain consistent with the forward integration process. This consistency is vital for stable and convergent training, avoiding potential issues arising from gradient approximation errors.

Furthermore, our method offers a more transparent representation of the model's dynamics, making it suitable for applications where interpretability is crucial. The polynomial coefficients provide explicit insights into the evolution of hidden states, enhancing our understanding of the underlying processes within the ODE solver.

Future work will focus on extending this approach to more complex and large-scale problems, as well as exploring further improvements in efficiency and interpretability. We also aim to generalize these principles to other machine learning algorithms beyond Neural ODEs. We believe this work represents a significant step forward in deep learning, with the potential to inspire new research directions and applications.

By bridging the gap between machine learning, numerical methods, and engineering, this research showcases the power of interdisciplinary approaches to tackle complex challenges. We hope our findings will foster further collaboration between these fields, leading to the development of even more efficient, interpretable, and powerful models.

**Appendix :**

**Theorem:** Let $\mathbf{k}_1$, $\mathbf{k}_2$, $\mathbf{k}_3$ and $\mathbf{k}_4$ be approximate derivative values given by equation (10), $\mathbf{f}(t\,, \mathbf{y}(t))$ is approximated over $[t_n, t_{n+1}]$, $t_{n+1} = t_n + h$, by $\mathbf{g}(t)$, the interpolating polynomial defined on *by*:

$$\mathbf{g}(t_n) = \mathbf{k}_1$$

$$\frac{\mathrm{d}\mathbf{g}}{\mathrm{d}t}(t_n) = \frac{1}{h}(-3\mathbf{k}_1 + 2(\mathbf{k}_2 + \mathbf{k}_3) - \mathbf{k}_4)$$

$$\mathbf{g}(t_n + \tfrac{h}{2}) = \tfrac{1}{2}(\mathbf{k}_2 + \mathbf{k}_3) \qquad \text{(A.1)}$$

$$\mathbf{g}(t_n + h) = \mathbf{k}_4$$

and we have:

$$\mathbf{g}(t) = \mathbf{f}\big(t\,, \mathbf{y}(t)\big) + \mathrm{O}(h^3) \qquad \text{(A.2)}$$

**Demonstration:**

We'll start by showing the relationships:

$$\mathbf{k}_1 = \mathbf{f}\big(t_n, \mathbf{y}(t_n)\big) + \mathrm{O}(h^3)$$

$$\frac{1}{h}(-3\mathbf{k}_1 + 2(\mathbf{k}_2 + \mathbf{k}_3) - \mathbf{k}_4) = \frac{\mathrm{d}\mathbf{f}\big(t_n, \mathbf{y}(t_n)\big)}{\mathrm{d}t} + \mathrm{O}(h^3)$$

$$\tfrac{1}{2}(\mathbf{k}_2 + \mathbf{k}_3) = \mathbf{f}\left(t_n + \tfrac{h}{2}, y\left(t_n + \tfrac{h}{2}\right)\right) + \mathrm{O}(h^3) \qquad \text{(A.1)}$$

$$\mathbf{k}_4 = \mathbf{f}\big(t_n + h, y(t_n + h)\big) + \mathrm{O}(h^3)$$

Denoting

$$\mathbf{f} = \mathbf{f}\big(t_n, \mathbf{y}(t_n)\big), \mathbf{f}'_t = \tfrac{\partial \mathbf{f}}{\partial t}\big(t_n, \mathbf{y}(t_n)\big), \mathbf{f}'_y = \tfrac{\partial \mathbf{f}}{\partial \mathbf{y}}\big(t_n, \mathbf{y}(t_n)\big), \mathbf{f}''_{t^2} = \tfrac{\partial^2 \mathbf{f}}{\partial t^2}\big(t_n, \mathbf{y}(t_n)\big), \mathbf{f}''_{ty} = \tfrac{\partial^2 \mathbf{f}}{\partial t \partial \mathbf{y}}\big(t_n, \mathbf{y}(t_n)\big), \mathbf{f}''_{z^2} = \tfrac{\partial^2 \mathbf{f}}{\partial \mathbf{y}^2}\big(t_n, \mathbf{y}(t_n)\big), \ldots \qquad \text{(A.2)}$$

Given $\varepsilon$ in [0,1], one can express $y(t_n + \varepsilon h)$ using a Taylor series expansion around $t_n$:

$$\mathbf{y}(t_n + \varepsilon h) = \mathbf{y}(t_n) + \varepsilon h\,\mathbf{f} + \tfrac{(\varepsilon h)^2}{2}\left(\mathbf{f}'_t + \mathbf{f}\mathbf{f}'_y\right) + \tfrac{(\varepsilon h)^3}{6}\left(\mathbf{f}^2\mathbf{f}''_{y^2} + 2\mathbf{f}\mathbf{f}''_{ty} + \mathbf{f}''_{t^2} + \left(\mathbf{f}'_t + \mathbf{f}\mathbf{f}'_y\right)\mathbf{f}'_y\right) + O(h^4) \qquad \text{(A.3)}$$

In the same way, one has:

$$\mathbf{f}(t_n + \varepsilon h, \mathbf{y}(t_n + \varepsilon h)) = \mathbf{f} + (\mathbf{f}'_t + \mathbf{f}\mathbf{f}'_z)\varepsilon h + \left(\mathbf{f}^2\mathbf{f}''_{z^2} + 2\mathbf{f}\mathbf{f}''_{tz} + \mathbf{f}''_{t^2} + (\mathbf{f}'_t + \mathbf{f}\mathbf{f}'_y)\mathbf{f}'_y\right)\frac{(\varepsilon h)^2}{2} + O(h^3) \qquad \text{(A.4)}$$

Development of intermediate slopes $\mathbf{k}_1$, $\mathbf{k}_2$, $\mathbf{k}_3$, and $\mathbf{k}_4$,

At the beginning of the step:

$$\mathbf{k}_1 = \mathbf{f}(t_n, \mathbf{y}(t_n))$$

For $\mathbf{k}_2$, we have

$$\mathbf{k}_2 = \mathbf{f}\left(t_n + \frac{h}{2}, \mathbf{y}_n + \frac{h}{2}\mathbf{k}_1\right)$$

Using Taylor's development of **f** around $(t_n, \mathbf{y}_n)$ up to $O(h^3)$:

$$\mathbf{k}_2 = \mathbf{f}\left(t_n + \frac{h}{2}, \mathbf{y}_n + \frac{h}{2}\mathbf{k}_1\right) = \mathbf{f}\left(t_n + \frac{h}{2}, \mathbf{y}_n + \frac{h}{2}\mathbf{f}\right)$$

$$\mathbf{k}_2 = \mathbf{f} + \frac{h}{2}(\mathbf{f}'_t + \mathbf{f}\mathbf{f}'_y) + \frac{1}{2}\left(\frac{h}{2}\right)^2\left(\mathbf{f}''_{t^2} + 2\mathbf{f}\mathbf{f}''_{ty} + \mathbf{f}^2\mathbf{f}''_{y^2}\right) + O(h^3)$$

what gives by using (A.4):

$$\mathbf{k}_2 = \mathbf{f}\left(t_n + \frac{h}{2}, \mathbf{y}\left(t_n + \frac{h}{2}\right)\right) + O(h^3)$$

we develop also $\mathbf{k}_3$,

$$\mathbf{k}_3 = \mathbf{f}\left(t_n + \frac{h}{2}, \mathbf{y}_n + \frac{h}{2}\mathbf{k}_2\right)$$

Using the Taylor expansion around $(t_n, \mathbf{y}_n)$:

$$\mathbf{k}_3 = \mathbf{f} + \frac{h}{2}(\mathbf{f}'_t + \mathbf{k}_2\mathbf{f}'_y) + \frac{1}{2}\left(\frac{h}{2}\right)^2\left(\mathbf{f}''_{t^2} + 2\mathbf{k}_2\mathbf{f}''_{ty} + {\mathbf{k}_2}^2\mathbf{f}''_{y^2}\right) + O(h^3)$$

Substituting the expansion of $\mathbf{k}_2$, neglecting terms superior to order $h^3$:

$$\mathbf{k}_3 = \mathbf{f} + \frac{h}{2}\left(\mathbf{f}'_t + \left(\mathbf{f} + \frac{h}{2}(\mathbf{f}'_t + \mathbf{f}\mathbf{f}'_y)\right)\mathbf{f}'_y\right) + \frac{1}{2}\left(\frac{h}{2}\right)^2\left(\mathbf{f}''_{t^2} + 2\mathbf{f}\mathbf{f}''_{ty} + \mathbf{f}^2\mathbf{f}''_{y^2}\right) + O(h^3)$$

$$\mathbf{k}_3 = \mathbf{f} + \frac{h}{2}\left(\mathbf{f}'_t + \mathbf{f}\mathbf{f}'_y\right) + \frac{1}{2}\left(\frac{h}{2}\right)^2 \left(\mathbf{f}''_{t^2} + 2\mathbf{f}\mathbf{f}''_{ty} + \mathbf{f}^2\mathbf{f}''_{y^2} + 2\left(\mathbf{f}'_t + \mathbf{f}\mathbf{f}'_y\right)\mathbf{f}'_y\right) + O(h^3)$$

And so, according to (A.4):

$$\mathbf{k}_3 = \mathbf{f}\left(t_n + \frac{h}{2}, \mathbf{y}\left(t_n + \frac{h}{2}\right)\right) + O(h^3)$$

Same for $\mathbf{k}_4$:

$$\mathbf{k}_4 = \mathbf{f}\left(t_n + \frac{h}{2}, \mathbf{y}_n + \frac{h}{2}\mathbf{k}_3\right)$$

Using the Taylor expansion around $(t_n, \mathbf{y}_n)$:

$$\mathbf{k}_4 = \mathbf{f} + h\left(\mathbf{f}'_t + \mathbf{k}_3\mathbf{f}'_y\right) + \frac{1}{2}h^2\left(\mathbf{f}''_{t^2} + 2\mathbf{k}_3\mathbf{f}''_{ty} + {\mathbf{k}_3}^2\mathbf{f}''_{y^2}\right) + O(h^3)$$

Substituting the expansion of $\mathbf{k}_2$, neglecting terms superior to order $h^3$:

$$\mathbf{k}_4 = \mathbf{f} + h\left(\mathbf{f}'_t + \left(\mathbf{f} + \frac{h}{2}\left(\mathbf{f}'_t + \mathbf{f}\mathbf{f}'_y\right)\right)\mathbf{f}'_y\right) + \frac{1}{2}h^2\left(\mathbf{f}''_{t^2} + 2\mathbf{f}\mathbf{f}''_{ty} + \mathbf{f}^2\mathbf{f}''_{y^2}\right) + O(h^3)$$

$$\mathbf{k}_4 = \mathbf{f} + h\left(\mathbf{f}'_t + \mathbf{f}\mathbf{f}'_y\right) + \frac{1}{2}h^2\left(\mathbf{f}''_{t^2} + 2\mathbf{f}\mathbf{f}''_{ty} + \mathbf{f}^2\mathbf{f}''_{y^2} + \left(\mathbf{f}'_t + \mathbf{f}\mathbf{f}'_y\right)\mathbf{f}'_y\right) + O(h^3)$$

And so, according to (A.4):

$$\mathbf{k}_4 = \mathbf{f}\left(t_n + h, \mathbf{y}(t_n + h)\right) + O(h^3)$$

As a summary:

$$\mathbf{k}_1 = \mathbf{f}$$

$$\mathbf{k}_2 = \mathbf{f} + \frac{h}{2}\left(\mathbf{f}'_t + \mathbf{f}\mathbf{f}'_y\right) + \frac{1}{2}\left(\frac{h}{2}\right)^2 \left(\mathbf{f}''_{t^2} + 2\mathbf{f}\mathbf{f}''_{ty} + \mathbf{f}^2\mathbf{f}''_{y^2}\right) + O(h^3)$$

$$\mathbf{k}_3 = \mathbf{f} + \frac{h}{2}\left(\mathbf{f}'_t + \mathbf{f}\mathbf{f}'_y\right) + \frac{1}{2}\left(\frac{h}{2}\right)^2 \left(\mathbf{f}''_{t^2} + 2\mathbf{f}\mathbf{f}''_{ty} + \mathbf{f}^2\mathbf{f}''_{y^2} + 2\left(\mathbf{f}'_t + \mathbf{f}\mathbf{f}'_y\right)\mathbf{f}'_y\right) + O(h^3)$$

$$\mathbf{k}_4 = \mathbf{f} + h\left(\mathbf{f}'_t + \mathbf{f}\mathbf{f}'_y\right) + \frac{1}{2}h^2\left(\mathbf{f}''_{t^2} + 2\mathbf{f}\mathbf{f}''_{ty} + \mathbf{f}^2\mathbf{f}''_{y^2} + \left(\mathbf{f}'_t + \mathbf{f}\mathbf{f}'_y\right)\mathbf{f}'_y\right) + O(h^3)$$

Furthermore, these latter relations show that:

$$-3\mathbf{k}_1 + 2(\mathbf{k}_2 + \mathbf{k}_3) - \mathbf{k}_4 = h\big(\mathbf{f}'_t + \mathbf{f}\mathbf{f}'_y\big) + O(h^3)$$

and since we have:

$$\frac{d\mathbf{f}}{dt}\big(t_n, y(t_n)\big) = \big(\mathbf{f}'_t + \mathbf{f}\mathbf{f}'_y\big)$$

We obtain:

$$\frac{1}{h}(-3\mathbf{k}_1 + 2(\mathbf{k}_2 + \mathbf{k}_3) - \mathbf{k}_4) = \frac{d\mathbf{f}}{dt}\big(t_n, \mathbf{y}(t_n)\big) + O(h^3)$$

The expression of $\mathbf{g}(t)$ on over $[t_n, t_{n+1}]$ can be written**:**

$$\mathbf{g}(t) = \mathbf{g}(t_n) + \frac{\mathrm{d}\mathbf{g}}{\mathrm{d}t}(t_n)\,(t - t_n) + \frac{\mathbf{a}_2}{h^2}(t - t_n)^2$$

or

$$\mathbf{g}(t) = \mathbf{a}_0 + \frac{\mathbf{a}_1}{h}(t_n)\,(t - t_n) + \frac{\mathbf{a}_2}{h^2}(t - t_n)^2$$

where

$\mathbf{a}_0 = \mathbf{k}_1$

$\mathbf{a}_1 = -3\,\mathbf{k}_1 + 2\,(\mathbf{k}_2 + \mathbf{k}_3) - \mathbf{k}_4$

$\mathbf{a}_2 = 2\,(\mathbf{k}_1 - (\mathbf{k}_2 + \mathbf{k}_3) + \mathbf{k}_4)$

As a result of interpolation of terms that approximate **f** to $O(h^3)$, we have

$$\mathbf{g}(t) = \mathbf{f}\big(t\,, \mathbf{y}(t)\big) + \mathrm{O}(h^3)$$

Furthermore, if $\tau = \frac{t - t_n}{h}$, then one has

$$\mathbf{g}(t) = \mathbf{p}(\tau) = \mathbf{a}_2\tau^2 + \mathbf{a}_1\tau + \mathbf{a}_0$$

By integration, we get also:

$$\mathbf{q}(\tau) = \mathbf{y}_n + \int_0^\tau h\mathbf{p}(\xi)d\xi = \mathbf{y}_n + \int_{t_n}^t \mathbf{g}(s)ds$$

which gives

$$\mathbf{q}(\tau) = \mathbf{y}_n + h\tau\left(\frac{1}{3}\mathbf{a}_2\tau^2 + \frac{1}{2}\mathbf{a}_1\tau + \mathbf{a}_0\right)$$

which satisfies for $\tau = 1$, the equation

$$\mathbf{q}(1) = \mathbf{y}_n + \frac{h}{6}(\mathbf{k}_1 + 2(\mathbf{k}_2 + \mathbf{k}_3) + \mathbf{k}_4)$$

so that, from Runge-Kutta

$\mathbf{q}(1) = \mathbf{y}(t_n+h) + O(h^5)$

As a result of these relations, we can estimate an approximation of $\mathbf{f}(t\,,\mathbf{y}(t))$ to $O(h^3)$ for all $t$ in $[t_n\,, t_{n+1}]$:

Put

$$\tau = \frac{t - t_n}{t_{n+1} - t_n} = \frac{t - t_n}{h}$$

$$\mathbf{p}(\tau) = \mathbf{g}(t) = \mathbf{a}_2\tau^2 + \mathbf{a}_1\tau + \mathbf{a}_0 = \mathbf{f}\big(t, \mathbf{y}(t)\big) + O(h^3)$$

$$\mathbf{q}(\tau) = \mathbf{y}_n + h\tau\left(\frac{1}{3}\mathbf{a}_2\tau^2 + \frac{1}{2}\mathbf{a}_1\tau + \mathbf{a}_0\right) = \mathbf{y}(t) + O(h^4)$$

And we can verify that

$$\mathbf{q}(1) = \mathbf{y}_n + \frac{\mathrm{h}}{6}(\mathbf{k}_1 + 2(\mathbf{k}_2 + \mathbf{k}_3) + \mathbf{k}_4)$$

So that

$$\mathbf{q}(1) = \mathbf{y}(\mathrm{t}_{n+1}) + \mathrm{O}(h^5)$$

**Error estimation:**

We can suppose the error constant over the interval $[t_n\,, t_{n+1}]$:

$$\mathbf{g}(t) = \mathbf{f}\big(t\,, \mathbf{y}(t)\big) + \boldsymbol{\varphi} h^3$$

So

$$\mathbf{g}(t_n + h) = \mathbf{f}\big(t_n + h\,, \mathbf{y}(t_n + h)\big) + \varphi h^3 = \mathbf{f}\big(t_n + h\,, \mathbf{q}(1) + \mathrm{O}(h^5)\big) + \boldsymbol{\varphi} h^3$$

And finally, neglecting O($h^5$) in front of $\boldsymbol{\varphi} h^3$,

$$\mathbf{g}(t_n + h) = \mathbf{f}(t_n + h\,, \mathbf{q}(1)) + \boldsymbol{\varphi} h^3$$

Which enables us to estimate the truncation error of approximation of $\mathbf{f}(t\,, \mathbf{y}(t))$ by $\mathbf{g}(t)$.

$$\boldsymbol{\varphi} h^3 = \mathbf{g}(t_n + h) - \mathbf{f}(t_n + h\,, \mathbf{q}(1))$$

Now consider $\boldsymbol{\phi}(t)$, the approximation of $\mathbf{y}(t)$ over $[t_n, t_{n+1}]$ given by:

$$\boldsymbol{\phi}(t) = \mathbf{y}_n + \int_{t_n}^{t} \mathbf{g}(s) ds$$

We have

$$\boldsymbol{\phi}(t) = \mathbf{y}_n + \int_{t_n}^{t} \mathbf{g}(s) ds = \mathbf{y}_n + \int_{t_n}^{t} \left(\mathbf{f}(s\,, \mathbf{y}(s)) + \boldsymbol{\varphi} h^3\right) ds = \mathbf{y}(t) + \boldsymbol{\varphi} h^3 (t - t_n)$$

This error is bounded by:

$$err = \mathrm{h}\left\|\mathbf{p}(1) - \mathbf{f}(t_n + h, \mathbf{q}(1))\right\|_\infty = \|\boldsymbol{\varphi}\|_\infty\, h^4$$

In particular, performing a step $h$ allows us to determine $\|\boldsymbol{\varphi}\|_\infty$, assumed constant:

$$\|\boldsymbol{\varphi}\|_\infty = \frac{err}{h^4}$$

And if we wish to obtain for the next step $h_1$ an error less than a given precision $\varepsilon$, which amounts to:

$$\|\boldsymbol{\varphi}\|_\infty\, h_1^4 = \frac{err}{h^4} h_1^4 < \varepsilon$$

you have to choose a size step

$$h_1 = S\, h \left(\frac{err}{\varepsilon}\right)_\infty^{-0.25}, \; S = 0.9$$